# Self-training for cross-centre coronary vessel segmentation in X-ray angiography using pre-contrast frame subtraction

Danial Moafi[1*], Aida Moafi[1*], Mostafa Aryafar[1], Ashkan Sami[2], Maryam Aryafar[1†]

*[1]APEC International Ltd, London, UK; [2]Edinburgh Napier University, Edinburgh, UK*

[*] Shared first authorship.

[†]Corresponding author: Maryam Aryafar (info@apec-international.com)

**Abstract**

Deep learning models for coronary vessel segmentation in X-ray angiography often perform poorly on data from centres that were not represented in training. Apart from differences in equipment, the labels of the largest public dataset (ARCADE) are incomplete, as only the segments defined in the SYNTAX score are annotated. In this study, we evaluated whether self-training with pseudo-labels can improve performance on external data without additional manual annotation. Pseudo-labels were generated from two sources, namely the predictions of an nnU-Net model trained on ARCADE and a contrast map obtained by subtracting a pre-contrast frame from the same sequence. The nnU-Net predictions were extended along the contrast map, and pixels detected only in the subtraction image were assigned an ignore label and excluded from the loss. Six models were trained and tested on ARCADE and on two external datasets (XCAD and XACV). All first-round models initialised with the baseline predictions performed better than both the baseline and a control model fine-tuned for the same number of epochs. The model trained on the XCAD pool improved the macro-averaged Dice score by 2.7 points (95% CI 2.3–3.2), while the control decreased it by 0.6 points. A second round of self-training did not lead to further improvement. When an external model trained on 50 fully annotated images was used to generate the initial labels, the XCAD Dice score increased from 0.623 to 0.745 without any XCAD labels, although performance on XACV did not improve because large main vessels were missed. These results suggest that the benefit of self-training depends largely on the sensitivity of the model used to generate the initial pseudo-labels.

## 1. Introduction

Quantitative coronary angiography (QCA) is used to assess the severity of coronary stenoses on invasive X-ray angiograms, and automated QCA requires accurate segmentation of the coronary arteries in each analysed frame. Deep learning methods such as nnU-Net (Isensee et al., 2021) perform well on this task when sufficient annotated data are available. However, annotated angiography datasets are few and relatively small. The largest public dataset, ARCADE (Popov et al., 2024), contains 1,000 annotated frames, and its annotations cover only the coronary segments defined in the SYNTAX score. Side branches and distal vessels outside the score are labelled as background.

Segmentation models trained on data from a single source often perform worse on images from other centres and vendors. For coronary angiography, external datasets with complete vessel annotations, such as XCAD (Ma et al., 2021) and XACV (Wu et al., 2025), allow this to be measured. When the training labels are incomplete, as in ARCADE, a model may also learn to leave unlabelled vessels unsegmented, which would further reduce sensitivity on fully annotated external data. Collecting complete annotations from additional centres is time-consuming and expensive, so methods that can make use of unlabelled images are of practical interest.

Several self-supervised and unsupervised methods have been proposed for vessel segmentation in angiography (Ma et al., 2021; Kim et al., 2023; Shi et al., 2023). Self-training with pseudo-labels (Lee, 2013; Xie et al., 2020) is a simpler alternative, in which a trained model is used to label new images and is then retrained on these labels. A known problem with this approach is confirmation bias, since the model is trained to repeat its own mistakes. One way to reduce this effect is to add a second, independent source of labels.

In coronary angiography, a second source of information is often available within the acquisition itself, because most sequences include frames recorded before the contrast agent arrives. Vesselness filters have previously been used to generate noisy training labels (Zhang et al., 2020), and Zeng et al. (2024) used learned single-frame subtraction to pretrain a segmentation model (DeepSA). To our knowledge, subtraction of a pre-contrast frame from the same sequence has not yet been used as an additional source of pseudo-labels in a self-training framework.

The aim of this study was to evaluate whether self-training with pseudo-labels from an nnU-Net model and from pre-contrast frame subtraction can improve coronary vessel segmentation on external datasets without new annotations. We also examined the effect of a second self-training round and of the choice of the initial model.

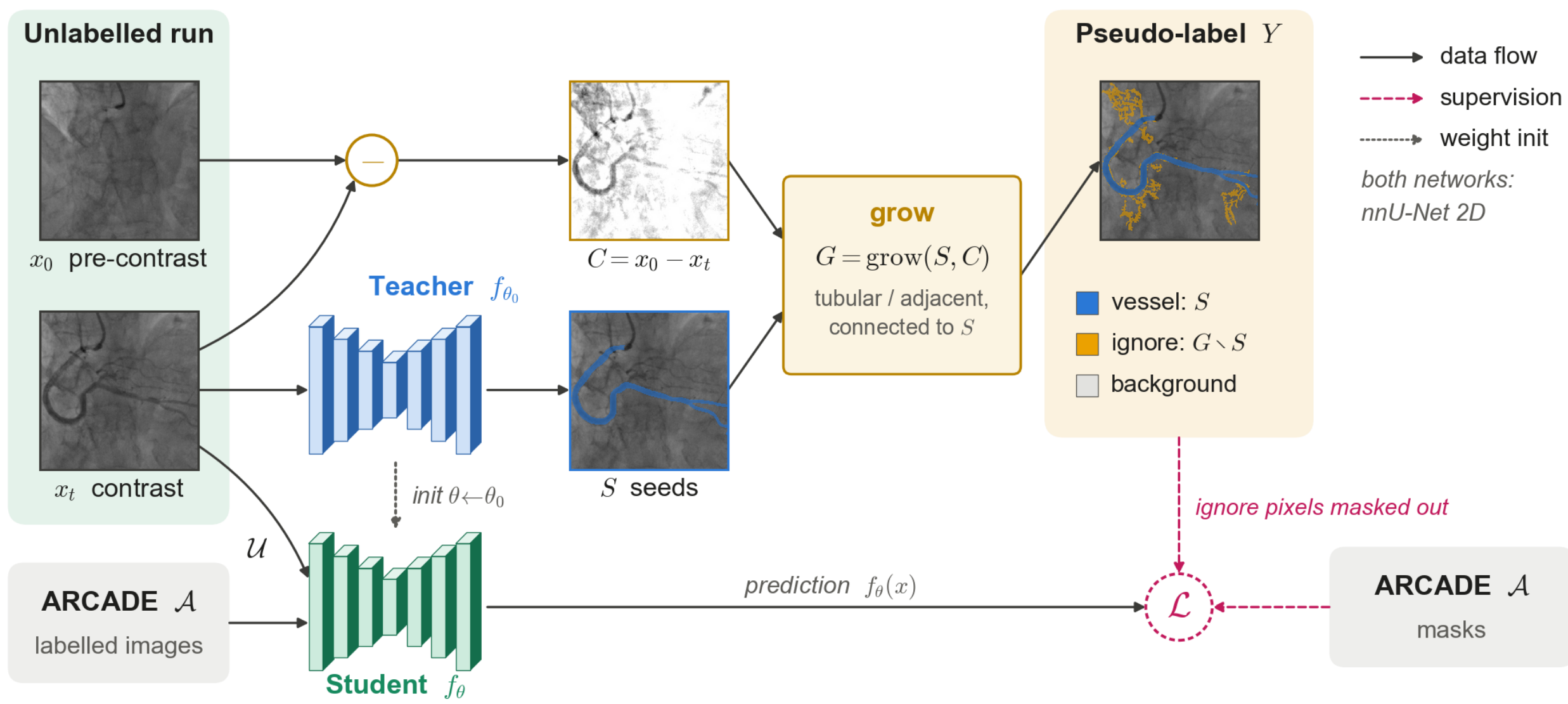

**Figure 1:** Overview of one self-training round. The baseline nnU-Net (teacher, $f_{\theta_0}$) segments the contrast-filled frame $x_t$ to give the seed mask S. The pre-contrast frame $x_0$ of the same sequence, after registration to $x_t$, is subtracted to give the contrast map C, along which S is extended by region growing (G). In the pseudo-label Y, pixels in S are labelled vessel (except low-contrast components, which are labelled ignore), pixels in G\S are labelled ignore, and all remaining pixels are labelled background. The student is initialised from the teacher weights and trained on the pseudo-labelled frames ($\mathcal{U}$) together with the labelled ARCADE images ($\mathcal{A}$), with ignore pixels excluded from the loss.

## 2. Method

An overview of the method is shown in Figure 1. In each self-training round, pseudo-labels were generated for a pool of unlabelled frames using two sources. The two labels were then combined into a three-class map (vessel, background and ignore), and the segmentation model was fine-tuned on these labels together with the original ARCADE training data. No manual annotations were added.

### 2.1 Pseudo-labels from the segmentation model

The first source is the baseline model, a 2D nnU-Net trained on the 1,000 ARCADE training frames for 311 epochs (training was stopped before the default 1,000-epoch schedule finished, and the checkpoint with the best validation score was used).

### 2.2 Pseudo-labels from frame subtraction

The second source is derived from the angiogram itself. Most angiographic sequences start before the contrast agent reaches the coronary arteries, and many continue until it has washed out. Subtracting such a pre-contrast frame from a later contrast-filled frame removes much of the static background and leaves mainly the contrast-filled vessels, similar to digital subtraction angiography. In this paper, a pre-contrast frame means any frame of the same sequence without coronary contrast, which can also be a frame after washout. For XCAD, the dataset provides such a background frame for each angiogram, matched by file name; one was available for 1,610 of the 1,621 training frames. For XACV, contrast and background frames were selected automatically. Each frame was given a contrast score, the mean Frangi dark-ridge response ($\sigma$ = 1, 2, 3 px; $\gamma$ = 0.05) on a copy downsampled to 256 × 256 px. The three highest-scoring frames of each sequence, at least three frames apart, were used as contrast frames, and each was paired with the nearest frame in time whose score lay in the lowest 15% of the range between the minimum and peak score of that sequence. This gave 252 frames from 84 sequences. For each frame in the pool, a pre-contrast frame from the same sequence was first registered to it using phase correlation (integer-pixel translation, maximum 20 px) to compensate for respiratory and table motion. The registered frame was then subtracted, and the difference image was normalised by its 99.5th percentile to obtain a contrast map (Figure 1). Because structures that moved between the two frames also appear in this map, a simple threshold does not give a reliable vessel mask. We therefore used the contrast map only to extend the nnU-Net prediction by region growing. A pixel adjacent to the current mask was added if its contrast value was above 0.15 and if it either had a Frangi vesselness response above $2\times10^{-3}$ ($\sigma$ = 1–6 px; Frangi et al., 1998) or lay within 3 px of the existing mask. The second condition was included because the vesselness response is low at vessel borders. Only regions connected to the original mask were kept. The contrast map was smoothed with a Gaussian filter ($\sigma$ = 1 px) before thresholding and before computing the vesselness response.

Several rules were applied to limit false positives. Added structures wider than 12 px were removed, since the vessels expected to be missed are thin distal branches, whereas structures such as the diaphragm border or contrast in the aortic root are wider. Frames with a nearly empty nnU-Net mask, and frames in which the added region covered more than 8% of the image, were excluded from the pool. In XACV, this removed 25 and 4 frames, respectively, leaving 223 of 252 frames for training. In addition, connected components of at least 30 px in the nnU-Net prediction with a mean contrast value below 0.05 were considered likely to be guidewires or collimator edges. These were labelled as ignore rather than background, because a real vessel may also show low contrast when some contrast agent is already present in the pre-contrast frame.

### 2.3 Label fusion and training

The two sources were combined pixel-wise into three classes. Pixels segmented by the nnU-Net model, except the low-contrast components described in Section 2.2, were labelled as *vessel*, pixels added only by region growing on the contrast map were labelled as *ignore*, and all remaining pixels were labelled as *background*. We used the ignore class because these pixels are the most likely to be vessels missed by the baseline model, and labelling them as background would reproduce the incomplete labelling of ARCADE.

The student model was trained using the `ignore_label` option in nnU-Net, which excludes ignore pixels from the loss. It was initialised from the baseline checkpoint and fine-tuned for 100 epochs on the ARCADE training set together with the pseudo-labelled pool.

### 2.4 Datasets and experiments

Table 1 summarises the datasets. All models were trained on the labelled ARCADE training set, and unlabelled frames from the XCAD and XACV training sets were used as pseudo-label pools. The test sets of all three datasets were used for evaluation.

**Table 1:** Datasets used in this study and their role. Numbers are frames; – indicates not used. No XCAD or XACV labels were used for training.

| Dataset | Centre / vendor | Annotation | Labelled training | Unlabelled pool (pseudo-labels) | Test |
|---|---|---|---|---|---|
| ARCADE (Popov et al., 2024) | Training centre | Partial (SYNTAX segments only) | 1,000 | – | 300 |
| XCAD (Ma et al., 2021) | Different vendor and centre | Complete | – | 1,621 (1,610 with a pre-contrast frame) | 126 |
| XACV (Wu et al., 2025) | Third centre | Complete | – | 252 (223 used) | 52 (14 cases) |

**Experiments.** Six models were compared. Two served as references, the baseline (Section 2.1) and a control model. The control was fine-tuned from the baseline for 100 epochs on ARCADE only, without pseudo-labels, to separate the effect of the pseudo-labels from that of longer training. The other four models were self-trained. Two were trained for one round, on the XCAD pool and on the XACV pool, respectively. The third was trained for a second round on the XCAD pool, with initial labels from the first-round XCAD model instead of the baseline. The fourth was trained on the XCAD pool with initial labels from DeepSA (Zeng et al., 2024), an external model trained on 50 fully annotated images that had not seen any XCAD or XACV labels. Each model was trained once, with a single random seed.

**Evaluation.** Dice and clDice (Shit et al., 2021) were calculated per image and averaged for each test set. Macro Dice was defined as the mean over the three test sets. 95% confidence intervals were estimated by paired percentile bootstrap against the baseline over test images (2,000 resamples).

**Table 2:** Dice and clDice for each test set, macro Dice, and the paired difference in macro Dice compared with the baseline (in points, with 95% confidence intervals). All confidence intervals in the last column exclude zero.

| Model | ARCADE | | XCAD | | XACV | | Macro Dice | Δ macro [95% CI] |
|---|---|---|---|---|---|---|---|---|
| | Dice | clDice | Dice | clDice | Dice | clDice | | |
| Baseline (ARCADE only) | 0.799 | 0.807 | 0.623 | 0.601 | 0.541 | 0.408 | 0.654 | – |
| Control (+100 ep ARCADE) | 0.792 | 0.795 | 0.628 | 0.606 | 0.526 | 0.395 | 0.648 | −0.6 [−1.1, −0.1] |
| XCAD r1 | 0.809 | 0.817 | 0.661 | 0.645 | 0.574 | 0.444 | 0.681 | +2.7 [+2.3, +3.2] |
| XCAD r2 | 0.799 | 0.804 | 0.660 | 0.638 | 0.574 | 0.435 | 0.678 | +2.4 [+1.9, +2.9] |
| XACV r1 | 0.805 | 0.811 | 0.648 | 0.628 | 0.574 | 0.436 | 0.675 | +2.1 [+1.7, +2.6] |
| XCAD, DeepSA initial labels | 0.797 | 0.805 | 0.745 | 0.788 | 0.546 | 0.415 | 0.696 | +4.2 [+3.5, +4.9] |

**Table 3:** Mean sensitivity / precision per test set.

| Model | ARCADE | XCAD | XACV |
|---|---|---|---|
| Baseline (ARCADE only) | 0.766 / 0.853 | 0.495 / 0.884 | 0.389 / 0.927 |
| Control (+100 ep ARCADE) | 0.752 / 0.853 | 0.505 / 0.875 | 0.369 / 0.947 |
| XCAD r1 | 0.789 / 0.846 | 0.546 / 0.871 | 0.423 / 0.924 |
| XCAD r2 | 0.773 / 0.845 | 0.544 / 0.869 | 0.422 / 0.925 |
| XACV r1 | 0.781 / 0.846 | 0.527 / 0.880 | 0.420 / 0.933 |
| XCAD, DeepSA initial labels | 0.765 / 0.849 | 0.715 / 0.793 | 0.393 / 0.934 |

## 3. Results

Table 2 shows the results. The baseline achieved Dice scores of 0.799 on ARCADE, 0.623 on XCAD and 0.541 on XACV. The control reduced macro Dice by 0.6 points (95% CI 0.1–1.1). Both first-round self-trained models performed better than the baseline and the control on all three test sets. The XCAD-pool model gave the largest gain. Its Dice increased by 3.8 points on XCAD and 3.3 points on XACV, and macro Dice by 2.7 points (95% CI 2.3–3.2). Sensitivity increased by 2 to 5 points across the test sets (Table 3). Across the self-trained models (excluding the model with DeepSA initial labels), precision decreased by at most 1.6 points.

The second XCAD round and the XACV-pool model did not improve on the first XCAD model. Both performed similarly or slightly worse on all test sets, with differences of up to 1.3 points.

With initial labels from DeepSA, XCAD Dice increased to 0.745 (+12.2 points, 95% CI 10.5–13.8) without using any XCAD labels. Sensitivity increased from 0.495 to 0.715, and precision decreased by 9 points (Table 3). This model also scored higher on XCAD than DeepSA itself, which reached a Dice of 0.69 when we evaluated it on the XCAD test set. On XACV, however, it had the lowest Dice of the self-trained models (0.546, compared with 0.541 for the baseline) and missed large main vessels (Figure 2).

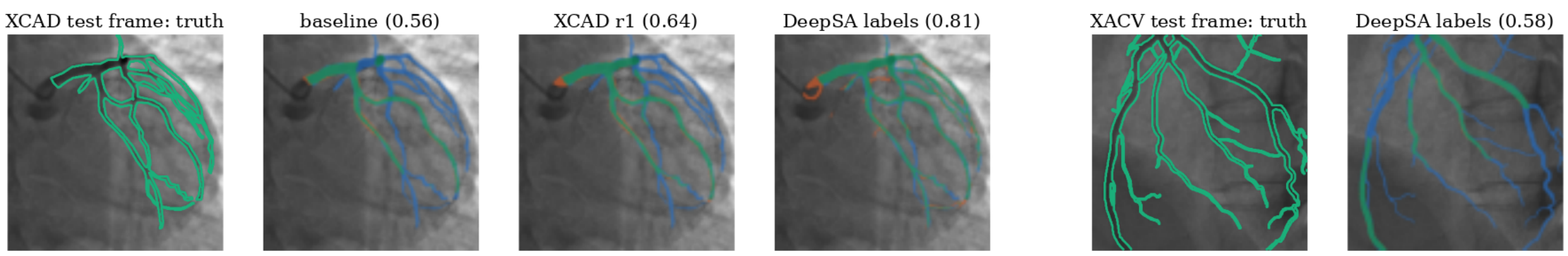

**Figure 2:** Left: segmentation of an XCAD test frame by the baseline model and two self-trained models (XCAD r1 and DeepSA initial labels). Right: an XACV test frame in which the model trained with DeepSA initial labels misses the main vessel. Green, true positive; blue, false negative; orange, false positive. Dice for the cropped region is given in brackets.

## 4. Discussion

Our results show that combining two sources of pseudo-labels with different types of error improved segmentation on external data, whereas longer training on ARCADE alone did not. Excluding uncertain pixels from the loss allowed the model to learn from regions where both sources agreed, without being trained on labels that were likely to be wrong. However, the improvement was limited by the sensitivity of the model used for the initial labels. The only experiment that clearly exceeded this limit on XCAD used a model trained on a small number of complete annotations. This suggests that the completeness of the labels may matter more than their number. For example, SMART (Luo et al., 2026) achieved a Dice of 0.844 on XACV using 16 labelled videos, compared with 0.574 for our best model trained without XACV labels.

The drop in baseline performance from ARCADE to the external datasets is probably not explained by differences in vendor and acquisition alone. The ARCADE test labels are incomplete in the same way as the training labels, so unsegmented distal vessels are not penalised, whereas XCAD and XACV are fully annotated and every missed branch is counted as an error. The pseudo-label statistics are consistent with this. On XACV, the baseline predictions covered 3.9% of image pixels, compared with 8.9% in the ground truth. This incomplete coverage probably also explains why a second round did not help. Region growing added only 1.7% of pixels on XACV, so much of the vessel tree remained labelled as background, and the first-round model missed many of the same vessels as the baseline.

The XCAD Dice of 0.745 obtained with DeepSA labels is higher than published results without XCAD labels, which range from 0.47 to 0.55 (Ma et al., 2021; Kim et al., 2023; Shi et al., 2023). The 0.755 reported for DeepSA itself was obtained by cross-validation on the XCAD test set and is therefore not directly comparable. The weaker result on XACV is probably related to the top-hat preprocessing in DeepSA, which suppresses wide dark structures such as large main vessels. The self-trained model inherited this error.

This study has some limitations. Each model was trained only once, so the confidence intervals reflect variability across test images but not across training runs. The subtraction step requires a suitable pre-contrast frame in the same sequence. This was available for 1,610 of 1,621 XCAD frames, but may not always be the case in clinical practice. The thresholds used for region growing were chosen by visual inspection of a small number of frames and were not tuned separately for each dataset. In addition, confidence intervals were computed over frames, and frames from the same patient are correlated, particularly in XACV (52 frames from 14 cases), so the intervals may be too narrow. Finally, the method was not evaluated on clinical data or on downstream QCA measurements.

In future work, we plan to fully annotate around 30 frames from a new clinical centre to obtain a better initial model for that setting. We also plan to explore self-supervised pretraining, either subtraction-based as in DeepSA or vessel-aware masked image modelling (Huang et al., 2026).

## 5. Conclusion

We showed that a pre-contrast frame from the same angiographic sequence can be used as a second source of pseudo-labels for self-training a coronary vessel segmentation model. One round of self-training improved performance on two external datasets without new annotations, but the size of the improvement was limited by the sensitivity of the initial model.

## Acknowledgements

This work was supported by Scottish Enterprise through a SMART: SCOTLAND Feasibility Grant awarded to APEC International Ltd.

## Data and code availability

ARCADE, XCAD and XACV are publicly available from their original publications. Code is available from the corresponding author on reasonable request.

## Ethics statement

This study used only publicly available, de-identified datasets, and no additional ethical approval was required.